\documentclass[runningheads]{llncs}
\usepackage{graphicx}
\usepackage[utf8]{inputenc}
\usepackage[breaklinks=true,bookmarks=false]{hyperref}
\usepackage{xcolor}
\usepackage[caption=false]{subfig}
\usepackage{multirow}
\usepackage{booktabs}
\usepackage{amsmath}

\begin{document}
\title{An efficient solution for semantic segmentation: ShuffleNet V2 with atrous separable convolutions} 
\titlerunning{An efficient solution for semantic segmentation}
%
\author{Sercan Türkmen\orcidID{0000-0002-3692-5019} \and Janne Heikkilä\orcidID{0000-0003-0073-0866}}
\authorrunning{S. Türkmen, J. Heikkilä}
%
\institute{Center for Machine Vision Research, University of Oulu, Oulu, Finland \email{sercanturkmen@outlook.com, janne.heikkila@oulu.fi}}
\maketitle              
\begin{abstract}

    Assigning a label to each pixel in an image, namely semantic segmentation, has been an important task in computer vision, and has applications in autonomous driving, robotic navigation, localization, and scene understanding. Fully convolutional neural networks have proved to be a successful solution for the task over the years but most of the work being done focuses primarily on accuracy. In this paper, we present a computationally efficient approach to semantic segmentation, while achieving a high mean intersection over union (mIOU), $70.33\%$ on Cityscapes challenge. The network proposed is capable of running real-time on mobile devices. In addition, we make our code and model weights publicly available.
    
\end{abstract}
\section{Introduction}

Semantic segmentation is a major challenge in computer vision that aims to assign a label to every pixel in an image.\cite{chen2018encoder,lin2014microsoft} Fully convolutional networks are shown to be the state-of-art approach in semantic segmentation tasks over the recent years and offer simplicity and speed during learning and inference\cite{long2015fully}. Such networks have a broad range of applications such as autonomous driving, robotic navigation, localization, and scene understanding.

These architectures are trained by supervised learning over numerous images and detailed annotations for each pixel. Data sets that offer semantic segmentation annotations on a rich variety of objects and stuff categories have emerged such as COCO\cite{lin2014microsoft}, ADE20K\cite{zhou2017scene}, Cityscapes\cite{cordts2016cityscapes}, PASCAL VOC\cite{everingham2015pascal}, thus opening new windows in the field.

Computationally efficient convolutional networks have been gaining momentum over the recent years but the segmentation task is still an open problem. Proposed networks for the semantic segmentation task are deep and resource hungry because of their purpose of achieving the highest accuracy such as \cite{chen2018searching,chen2018encoder,zhao2017pyramid,ronneberger2015u}. These approaches have high complexity and may contain custom operations which are not suitable to be run on current implementations of neural network interpreters offered for mobile devices. Such devices lack the computation power of specialised GPUs, resulting in very poor inference speed. Mobile capable approaches in the feature extraction task such as Mobilenet V2\cite{sandler2018mobilenetv2}, ShuffleNet V2\cite{ma2018shufflenet} have motivated us to explore the performance of such architectures to be used in this context.

In this paper, we explore ShuffleNet V2\cite{ma2018shufflenet}, as the feature extractor with simplified DeepLabV3+\cite{chen2018encoder} heads and recently proposed DPC\cite{chen2018searching} architecture, and report our findings of both model on scene understanding using Cityscapes\cite{cordts2016cityscapes} data set. Furthermore, we present the number of floating point operations(FLOPs) and on-device inference performance of each approach. Our contributions to the field can be listed in three points:

\begin{enumerate}
    \item We achieve state-of-art computation efficiency in the semantic segmentation task while achieving $70.33\%$ mean intersection over union (mIOU) on Cityscapes test set using ShuffleNet V2 along with DPC\cite{chen2018searching} encoder and a naive decoder module.
    \item Our proposed model and implementation is fully compatible with TensorFlow Lite and runs real-time on Android and iOS-based mobile phones.
    \item We make our Tensorflow implementation of the network, and trained models publicly available at \url{https://github.com/sercant/mobile-segmentation}\footnote{DOI: \url{https://doi.org/10.5281/zenodo.2620377}}.
\end{enumerate}

\section{Related Work}

In this section, we talk about the current state-of-art in the task of semantic segmentation, especially mobile capable approaches and performance metrics to measure the efficiency of networks.

CNNs have shown to be the state-of-art method for the task of semantic segmentation over the recent years. Especially fully convolutional neural networks (FCNNs) have demonstrated great performance on feature generation task and end-to-end training and hence is widely used in semantic segmentation as encoders. Moreover, memory friendly and computationally light designs such as \cite{howard2017mobilenets,sandler2018mobilenetv2,zhang2017shuffle,ma2018shufflenet}, have shown to perform well in speed-accuracy trade-off by taking advantage of approaches such as depthwise separable convolution, bottleneck design and batch normalization\cite{ioffe2015batch}. These efficient designs are promising for usage on mobile CPUs and GPUs, hence motivated us to use such networks as encoders for the challenging task of semantic segmentation.

FCNN models for semantic segmentation proposed in the field have been the top-performing approach in many benchmarks such as \cite{lin2014microsoft,cordts2016cityscapes,everingham2015pascal,zhou2017scene}. But these approaches use deep feature generators and complex reconstruction methods for the task, thus making them unsuitable for mobile use, especially for the application of autonomous cars where resources are scarce and computation delays are undesired \cite{siam2018comparative}.

In this sense, one of the recent proposals in feature generation, ShuffleNet V2\cite{ma2018shufflenet}, demonstrates significant efficiency boost over the others while performing accurately. According to \cite{ma2018shufflenet}, there are four main guidelines to follow for achieving a highly efficient network design.

\begin{enumerate}
    \item When the channel widths are not equal, there is an increase in the memory access cost (MAC) and thus, channel widths should be kept equal.
    \item Excessive use of group convolutions should be avoided as they raise the MAC.
    \item Fragmentation in the network should be avoided to keep the degree of parallelism high.
    \item Element-wise operations such as ReLU, Add, AddBias are non-negligible and should be reduced.
\end{enumerate}

To achieve such accuracy at low computation latency, they point out two main reasons. First, their guidelines for efficiency allow each building block to use more feature channels and have a bigger network capacity. Second, they achieve a kind of ``feature reuse" by their approach of keeping half of the feature channels pass through the block to join the next block.

Another important issue is the metric of performance for convolutional neural networks. The efficiency of CNNs is commonly reported by the total number of floating point operations (FLOPs). It is pointed out in \cite{ma2018shufflenet} that, despite their similar number of FLOPs, networks may have different inference speeds, emphasizing that this metric alone can be misleading and may lead to poor designs. They argue that discrepancy can be due to memory access cost (MAC), parallelism capability of the design and platform dependent optimizations on specific operations such as cuDNN’s $ 3 \times 3 $ Conv. Furthermore, they offer to use a direct metric (e.g., speed) instead of an indirect metric such as FLOPs.

Next, we shall examine the state-of-art on the semantic image segmentation task by focusing on the computationally efficient approaches.

\begin{figure}[t]
    \centering
    \subfloat[$rate=1 \times 1$]{\includegraphics[width=.21\linewidth]{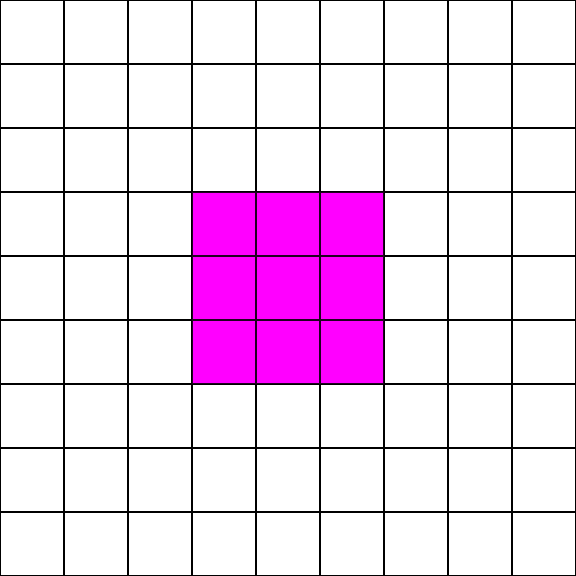}}\hfill
    \subfloat[$rate=2 \times 2$]{\includegraphics[width=.21\linewidth]{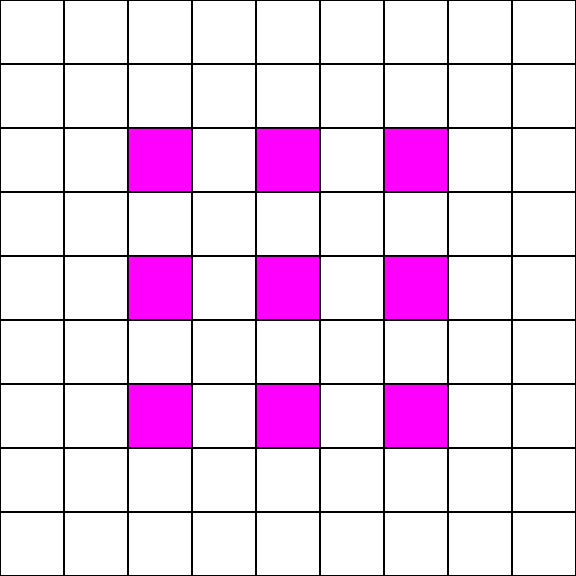}}\hfill
    \subfloat[$rate=3 \times 2$]{\includegraphics[width=.21\linewidth]{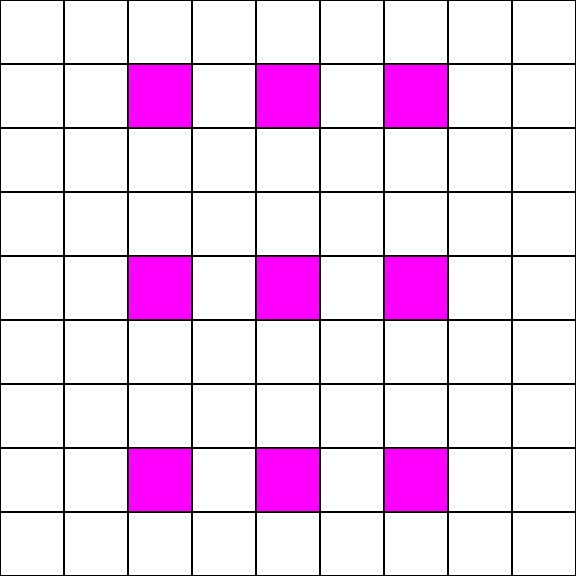}}\hfill
    \subfloat[$rate=2 \times 4$]{\includegraphics[width=.21\linewidth]{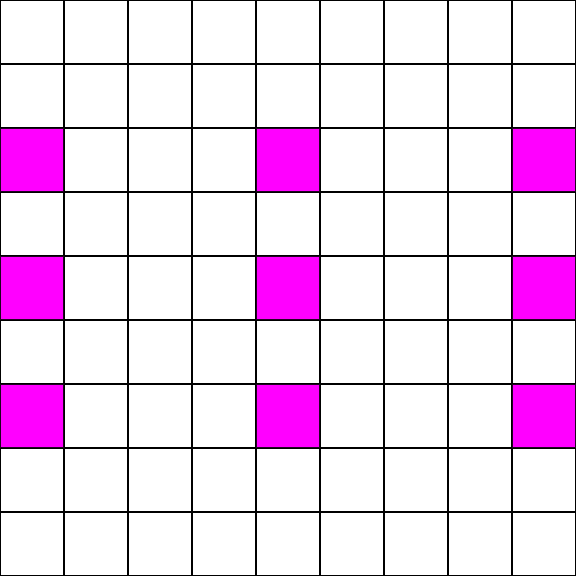}}
    \caption{Atrous convolutions, also known as dilated convolutions, with a kernel size of $3 \times 3$ at different dilation rates.}
    \label{fig:atrous}
\end{figure}

Atrous convolutions, or dilated convolutions, are shown to be a powerful tool in the semantic segmentation task \cite{chen2017rethinking}. By using atrous convolutions it is possible to use pretrained ImageNet networks such as \cite{ma2018shufflenet,sandler2018mobilenetv2} to extract denser feature maps by replacing downscaling at the last layers with atrous rates, thus allowing us to control the dimensions of the features. Furthermore, they can be used to enlarge the field of view of the filters to embody multi-scale context. Examples of atrous convolutions at different rates are shown in Figure \ref{fig:atrous}.

DeepLabV3+ DPC \cite{chen2018searching} achieves state-of-art accuracy when it is combined with their modified version of Xception\cite{chollet2017xception} backbone. In their work, Mobilenet V2 has shown to have a correlation of accuracy with Xception\cite{chollet2017xception} while having a shorter training time, and thus it is used in the random search\cite{bergstra2012RandomSF} of a dense prediction cell (DPC). Our work is inspired by the accuracy that they have achieved with the Mobilenet V2 backbone on Cityscapes set in \cite{chen2018searching}, and their approach of combining atrous separable convolutions with spatial pyramid pooling in \cite{chen2017rethinking}. To be more specific, we use lightweight prediction cell (denoted as basic) and DPC which were used on the Mobilenet V2 features, and the atrous separable convolutions on the bottom layers of feature extractor in order to keep higher resolution features.

Semantic segmentation as a real-time task has gained momentum on popularity recently. ENet \cite{paszke2016enet} is an efficient and lightweight network offering low number of FLOPs in the design and ability to run real-time on NVIDIA TX1 by taking advantage of bottleneck module. Recently, ENet was further fine-tuned by \cite{berman2018lovasz}, increasing the Cityscapes mean intersection over union from 58.29\% to 63.06\% by using a new loss function called Lovasz-Softmax \cite{berman2018lovasz}. Furthermore, SHUFFLESEG\cite{gamal2018shuffleseg}, demonstrates different decoders that can be used for Shufflenet, prior work to ShuffleNet V2, comparing their efficiency mainly with ENet\cite{paszke2016enet} and SegNet\cite{badrinarayanan2015segnet} by FLOPs and mIOU metrics but they do not mention any direct speed metric that is suggested by ShuffleNet V2\cite{ma2018shufflenet}. The most comprehensive work on the search for an efficient real-time network was done in \cite{siam2018comparative} and they report that SkipNet-Shufflenet combination runs $15$ frames per second (fps) with an image resolution of $ 640 \times 360 $ on Jetson TX2. This work again, like SHUFFLESEG\cite{gamal2018shuffleseg}, is based on the prior design of the channel shuffle based approach.

Our literature review showed us that the ShuffleNet V2 architecture is yet to be used in semantic segmentation task as a feature generator. Both \cite{siam2018comparative} and SHUFFLESEG\cite{gamal2018shuffleseg} point out the low FLOP achievable by using ShuffleNet and show comparable accuracy and fast inference speeds. In this work, we exploit improved ShuffleNet V2 as an encoder module modified by atrous convolutions and well-proven encoder heads of DeepLabV3 and DPC in conjunction. Then, we evaluate the network on Cityscapes, a challenging task in the field of scene parsing.

\section{Methodology}

This section describes our network architecture, and the training procedures, and evaluation methods. The network architecture is based upon the state-of-art efficient encoder, ShuffleNet V2\cite{ma2018shufflenet}, and DeepLabV3\cite{chen2017rethinking} and DPC\cite{chen2018searching} heads built on top to perform segmentation. Training procedure includes restoring from an Imagenet\cite{imagenet_cvpr09} checkpoint, pre-training on MS COCO 2017\cite{lin2014microsoft} and Cityscapes\cite{cordts2016cityscapes} coarse annotations, and fine-tuning on fine annotations of Cityscapes\cite{cordts2016cityscapes}. We then evaluate the trained network on Cityscapes validation set according to their evaluation procedure in \cite{cordts2016cityscapes}.

\subsection{Network Architecture}

\begin{figure}[t]
    \centering
    \includegraphics[width=\linewidth]{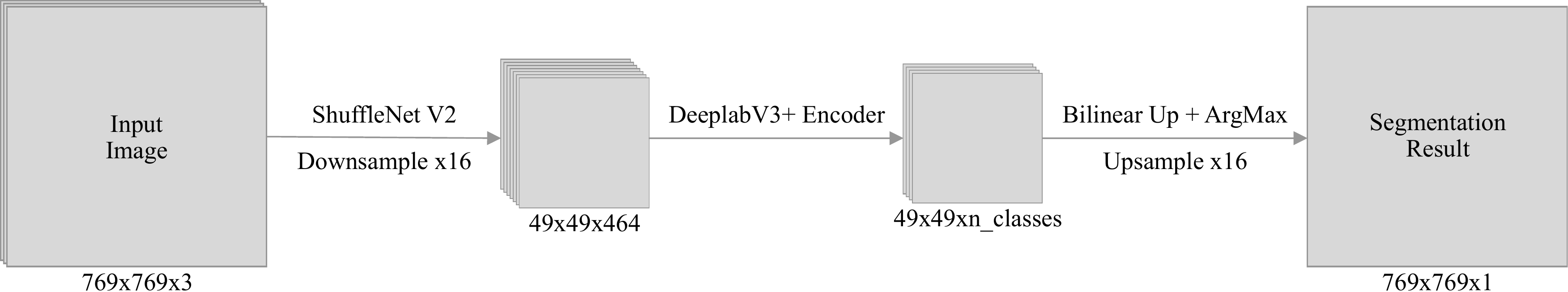}
    \caption{General view of the network at 769x769 with $output\_stride=16$.}
    \label{fig:general_flow}
\end{figure}

In this section, we will give a detailed explanation of each step of the proposed network. Figure \ref{fig:general_flow} shows the different stages of the network and how the data flows from the start to the end. We start with Shufflenet V2 feature extractor, then add the encoder head (DPC or basic DeepLabV3), and finally use resize bilinear as a naive decoder to produce the segmentation mask. The final downsampling factor of the feature extractor is denoted as $output\_stride$.

For the task of feature extraction, we choose Shufflenet V2 because of its success for speed versus accuracy trade-off. Our selection for the $depth\_multiplier$ of ShuffleNet V2 is $\times 1$ as it can be seen in the output channels column of Table \ref{tbl:network}. Our decision was purely made by accuracy and speed results of this variation on the ImageNet data set and we have not run experiments on different variations of the $depth\_multiplier$. Although, one might choose lower values for this hyperparameter to achieve faster inference time by compromising on accuracy and conversely higher values might result in a gain of accuracy in favour of inference speed.

\begin{figure}[t]
    \centering
    \subfloat[Basic unit]{\includegraphics[height=.7\linewidth]{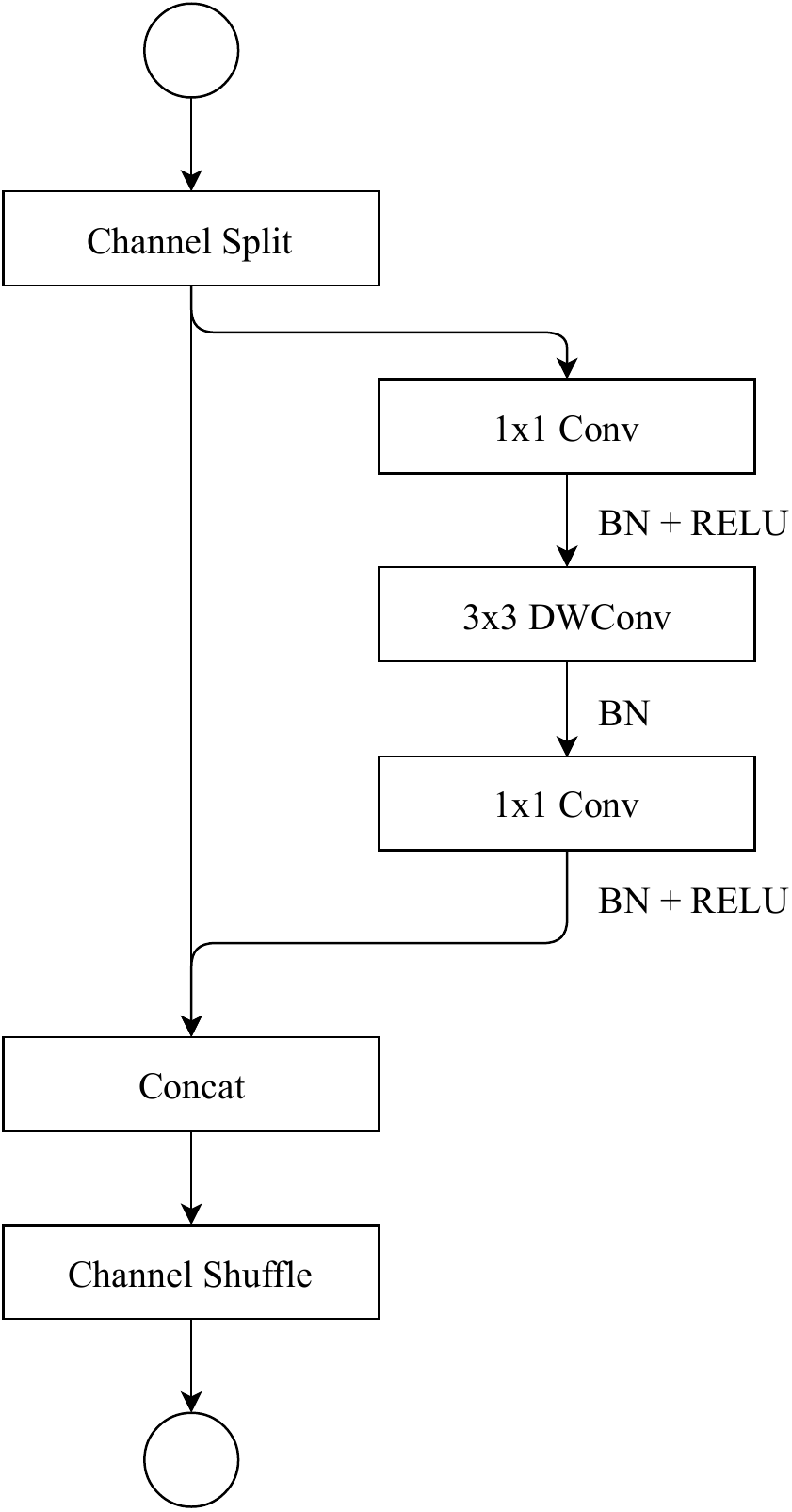}}\hfill
    \subfloat[Spatial downsampling unit]{\includegraphics[height=.7\linewidth]{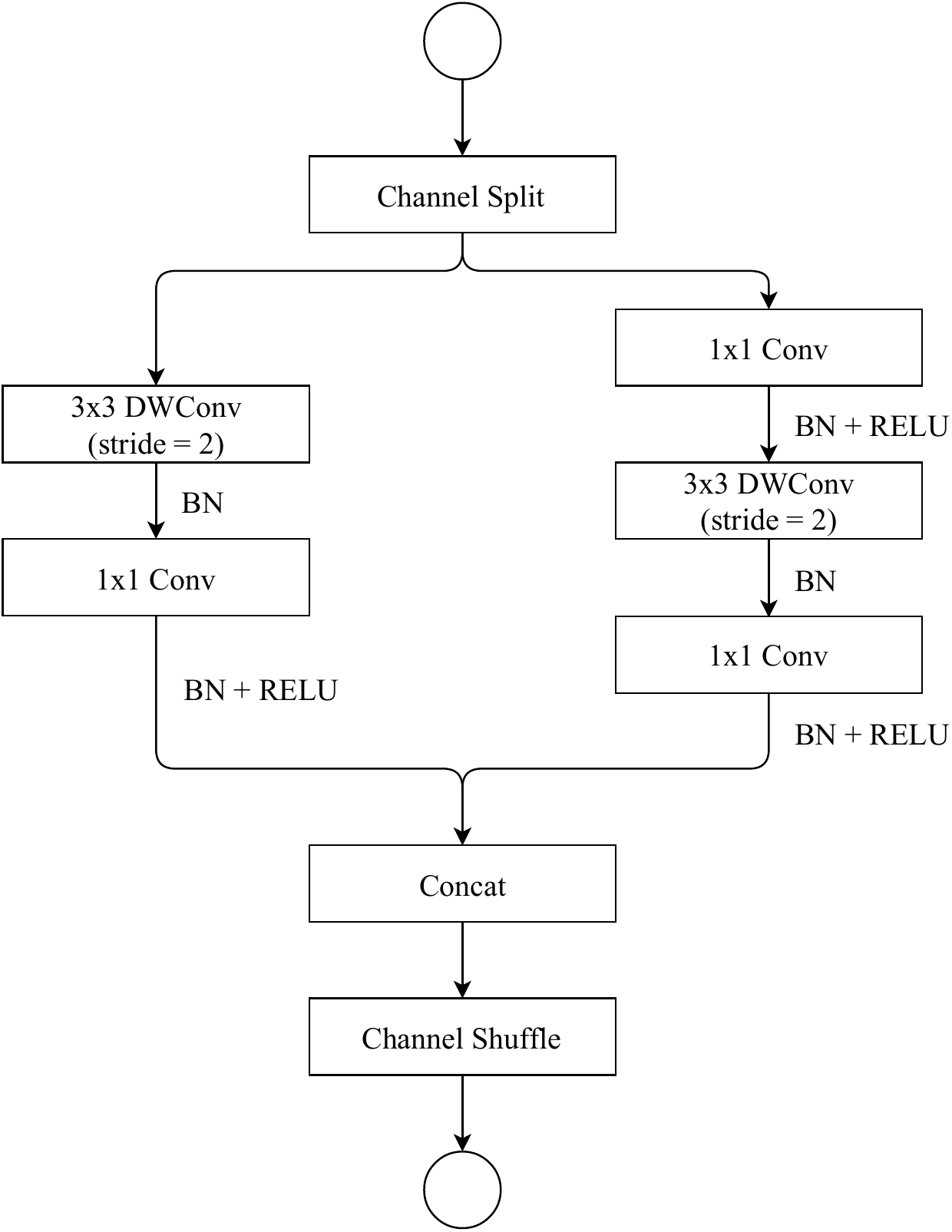}}
    \caption{Shufflenet V2 units (\textbf{DWConv} means depthwise convolution).}
    \label{fig:shufflenet}
\end{figure}

Figure \ref{fig:shufflenet} shows the building blocks of the feature extractor architecture in detail. Each stage consists of one spatial downsampling unit and several basic units. In the original implementation of Shufflenet V2\cite{ma2018shufflenet}, $output\_stride$ goes as low as $32$. In our approach, entry flow, $Stage2$ and $Stage3$ of the proposed feature extractor is implemented as in \cite{ma2018shufflenet}. In the case of $output\_stride=16$, the last stage, namely $Stage4$, has been modified by setting $stride=1$ instead of $2$ on the downsampling layer and the atrous rate of the preceding depth-wise convolutions are set to $network\_stride$ divided by $output\_stride$ as described in \cite{chen2017rethinking} to adjust the final downsampling factor of the feature extractor. However, in the case of $output\_stride=8$, stride and atrous rate modification starts from $Stage3$. We choose to focus on $output\_stride=16$ due to its faster computation speed.

\begin{figure}[t]
    \centering
    \subfloat[Basic encoder head.]{\includegraphics[height=.51\linewidth]{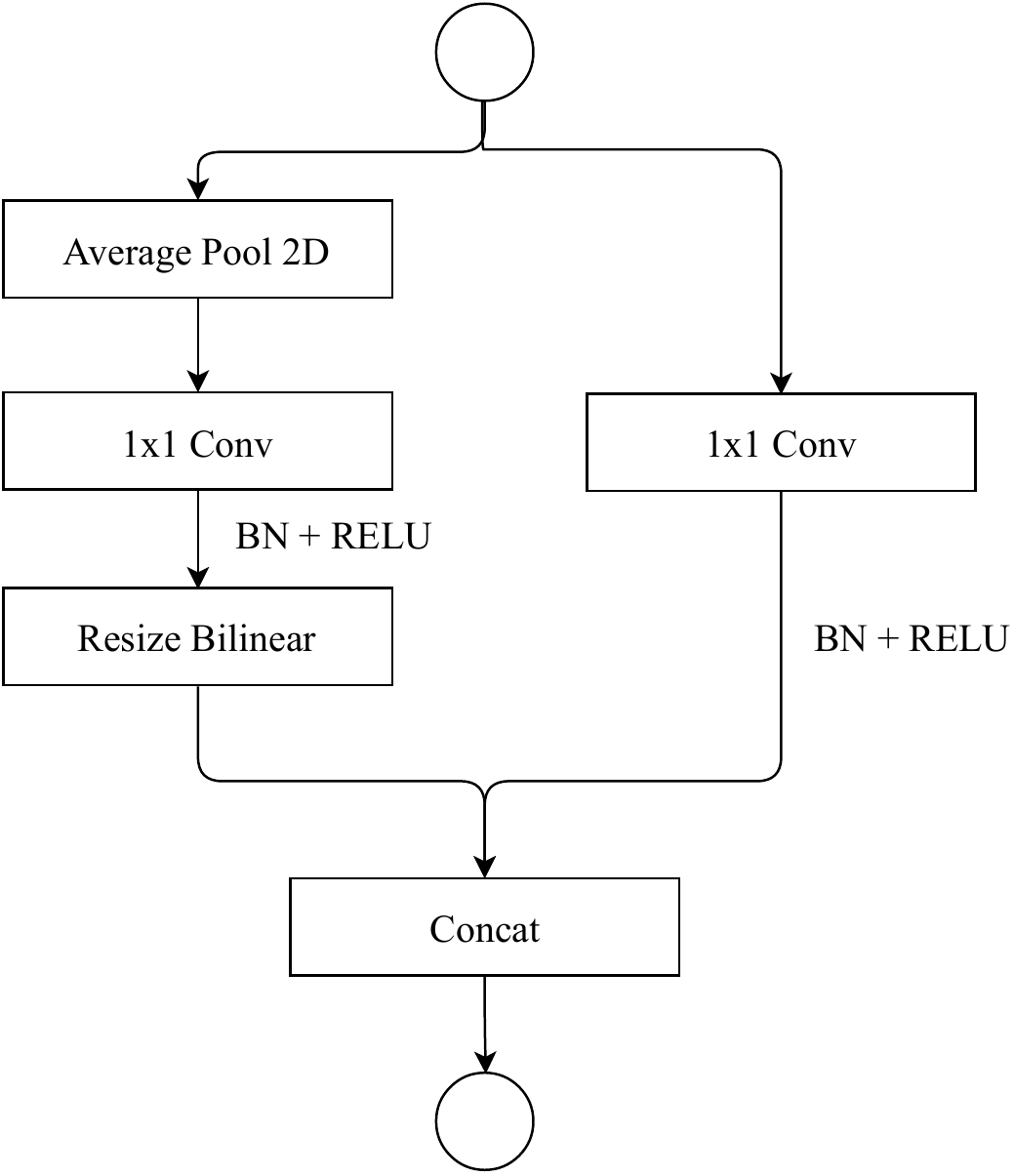}}\hfill
    \subfloat[DPC encoder head.]{\includegraphics[height=.51\linewidth]{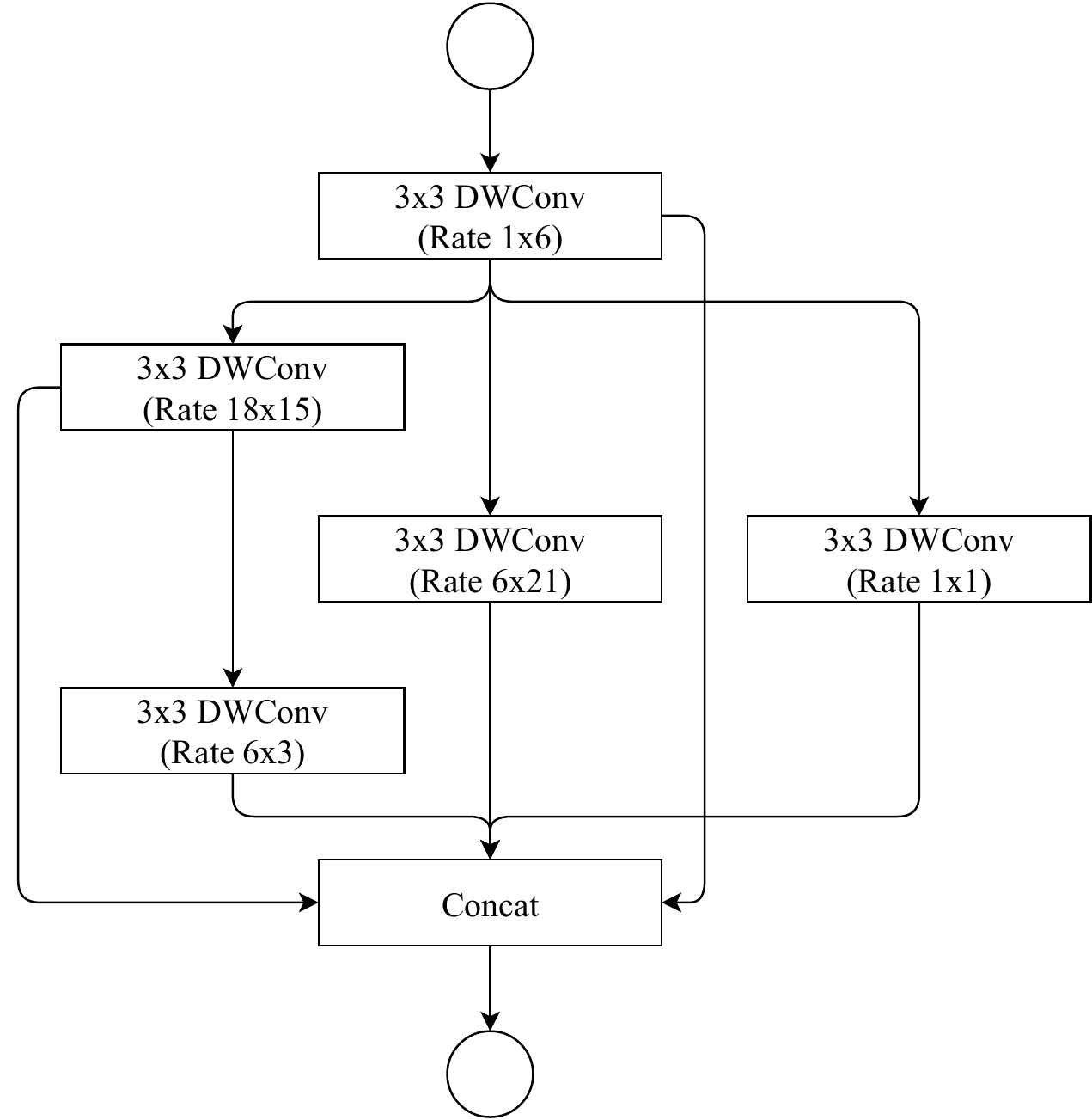}}
    \caption{Encoder head variants.}
    \label{fig:deeplabv3encoder}
\end{figure}

\begin{figure}
    \centering
    \includegraphics[width=\linewidth]{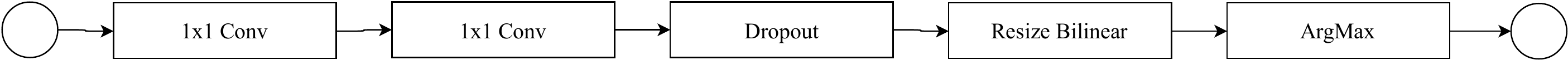}
    \caption{Exit flow.}
    \label{fig:exit_flow}
\end{figure}

After ShuffleNet V2 features are extracted we employ the DPC encoder. Figure \ref{fig:deeplabv3encoder} shows the design of the basic encoder head\cite{chen2017rethinking} and the DPC head\cite{chen2018searching}. The basic encoder head does not contain any atrous convolutions in its design for lower complexity. On the other hand, DPC is using five different depthwise convolutions at different rates to understand features better. After the encoder heads, as seen in Figure \ref{fig:exit_flow}, features are reduced down to $depth=256$, then to the number of classes. Afterwards, a drop out layer with 0.9 probability of keeping is applied.

For decoding, we are using the same naive approach as \cite{chen2017rethinking}, where a simple bilinear resizing is applied to upscale back to the original image dimensions. In the case of $output\_stride=16$, the upscaling factor is 16. In their later work, DeepLabV3+\cite{chen2018encoder}, they propose an approach where features from earlier stages (e.g. $decoder\_stride$ of $4$) are concatenated with the upsampled features of the last layer where the upscaling factor is $output\_stride$ divided by $decoder\_stride$ to preserve the finer details in the segmentation result. We have not included this part as it adds more complexity and slowing down the inference time.

\begin{table}[t]
    \centering
    \caption{Proposed network architecture at $output\_stride=16$. \textbf{DLV3} denotes DeepLabV3+.}
    \label{tbl:network}
    \begin{tabular}{c l c c c c c c c}
                                                                                         & Layer                    & Output Size        & Kernel         & Stride & Rate & Repeat             & Output Channels      \\
        \toprule
                                                                                         & Image                    & $ 769 \times 769 $ &                &        &      &                    & 3                    \\
        \cline{1-8}
        \parbox[t]{4mm}{\multirow{8}{*}{\rotatebox[origin=c]{90}{\emph{ShuffleNet V2}}}} & Conv2D                   & $ 385 \times 385 $ & $ 3 \times 3 $ & 2      &      & \multirow{2}{*}{1} & \multirow{2}{*}{24}  \\
                                                                                         & MaxPool                  & $ 193 \times 193 $ & $ 3 \times 3 $ & 2      &      &                    &                      \\
        \cline{2-8}
                                                                                         & \multirow{2}{*}{Stage 2} & $ 97 \times 97 $   &                & 2      & 1    & 1                  & \multirow{2}{*}{116} \\
                                                                                         &                          & $ 97 \times 97 $   &                & 1      & 1    & 3                  &                      \\
        \cline{2-8}
                                                                                         & \multirow{2}{*}{Stage 3} & $ 49 \times 49 $   &                & 2      & 1    & 1                  & \multirow{2}{*}{232} \\
                                                                                         &                          & $ 49 \times 49 $   &                & 1      & 1    & 7                  &                      \\
        \cline{2-8}
                                                                                         & \multirow{2}{*}{Stage 4} & $ 49 \times 49 $   &                & 1      & 1    & 1                  & \multirow{2}{*}{464} \\
                                                                                         &                          & $ 49 \times 49 $   &                & 1      & 2    & 3                  &                      \\
        \cline{1-8}
        \parbox[t]{4mm}{\multirow{3}{*}{\rotatebox[origin=c]{90}{\emph{DLV3}}}}          & DPC                      & $ 49 \times 49 $   &                & 1      & 1    & 1                  & 512                  \\
        \cline{2-8}
                                                                                         & Conv2D                   & $ 49 \times 49 $   & $ 1 \times 1 $ & 1      &      &                    & 256                  \\
                                                                                         & Conv2D                   & $ 49 \times 49 $   & $ 1 \times 1 $ & 1      &      &                    & $n\_classes$         \\
        \cline{1-8}
                                                                                         & Bilinear Up              & $ 769 \times 769 $ &                &        &      &                    & $n\_classes$         \\
                                                                                         & ArgMax                   & $ 769 \times 769 $ &                &        &      &                    & 1                    \\
        \bottomrule
    \end{tabular}
\end{table}

\subsection{Training}

Proposed feature extractor weights are initially restored from an ImageNet\cite{imagenet_cvpr09} checkpoint. After that, we pre-train the network end-to-end on MS COCO\cite{lin2014microsoft}. Further pre-training was done on Cityscapes coarse annotations before we finalize our training on Cityscapes fine annotations. During the training at each step data augmentation was performed on the images by randomly scaling between $0.5$ and $2.0$ with steps of $0.25$, randomly cropping by $ 769 \times 769 $ and randomly flipping left and right.

\subsubsection{Preprocessing}
For preprocessing the input images, we standardize each pixel to $[-1, 1]$ range according to Equation \ref{eq:zero_mean_unit_range}.

\begin{equation}
    inputs \times \frac{2}{255} - 1
    \label{eq:zero_mean_unit_range}
\end{equation}

\subsubsection{MS COCO 2017}

We pre-train the network on MS COCO as suggested in \cite{chen2018searching,chen2018encoder,chen2017rethinking}. Only the relevant classes to the Cityscapes task, which are person, car, truck, bus, train, motorcycle, bicycle, stop sign and parking meter, were chosen and other classes were marked as background. Also, we further filter the samples by criteria of having a non-background class area of over $1000$ pixels. This yields us $69795$ training and $2956$ validation samples. Training was done end-to-end by using a batch size of $16$, $output\_stride$ of $16$, weight decay set to $4e-5$ to prevent over-fitting, and a ``poly" learning rate policy as shown in Equation \ref{eq:poly} where $lr^{(k)}$ is the learning rate at step $k$, $lr_{initial}$ set to $0.001$ and $power$ set to $0.9$\cite{chen2017rethinking} for 60K steps using Adam optimizer\cite{kingma2014adam}($beta1=0.9$, $a2=0.999$, $epsilon=10^{-8}$).

\begin{equation}
    lr^{(k)} = lr_{initial} \left( 1 - \dfrac{k}{max\_iter} \right)^{power}
    \label{eq:poly}
\end{equation}

\subsubsection{Cityscapes}
First, we pre-train the network further using 20K coarsely annotated samples for 60K steps, following the same parameters that we used in MS COCO training.
Then, as the final step, we fine-tune the network for 120K steps on finely annotated set ($2975$ train, $500$ validation samples) with an initial learning rate of $0.0001$, a slow start at learning rate $1e-5$ for the first 186 steps. We found out that lowering the learning rate at the fine-tuning stage is crucial to achieving good accuracy. Other parameters are kept the same as the previous steps.

\begin{table}[t]
    \centering
    \caption{Cityscapes \textbf{validation} set performance. GFLOPs is measured on image resolution of $ 640 \times 360 \times 3 $.}
    \label{tbl:mIOU_val}
    \begin{tabular}{lcc}
        Method                                       & GFLOPs & mIOU$(\%)$ \\
        \toprule        
        SkipNet-ShuffleNet\cite{siam2018comparative} & 2.0    & 55.5       \\
        ENet\cite{siam2018comparative}               & 3.83   & n/a        \\
        MobileNet V2  $+$ Basic                      & 4.69   & 70.7       \\
        MobileNet V2  $+$ DPC                        & 5.56   & n/a        \\
        \hline
        ShuffleNet V2 $+$ Basic (Ours)               & 2.18   & 67.7       \\
        ShuffleNet V2 $+$ DPC (Ours)                 & 3.05   & 71.3       \\
        \bottomrule
    \end{tabular}
\end{table}

\section{Experimental Results on Cityscapes}

Cityscapes\cite{cordts2016cityscapes} is a well-studied data set in the field of scene parsing task. It contains roughly 20K coarsely annotated samples, and 5000 finely annotated samples which are split into 2975, 500 and 1525 for training, validation, and testing respectively.

Our approach of combining ShuffleNet V2 with DeepLabV3 basic and DPC head achieves both state-of-art mIOU and inference speed with the respective GFLOPs counts. Table \ref{tbl:mIOU_val} shows the comparison of GFLOPs to mIOU performance on the validation set. ShuffleNet V2 with basic DeepLabV3 encoder head, having $2.18$ GFLOPs, surpasses the SkipNet-ShuffleNet\cite{siam2018comparative} which has similar floating operations count by $12.2\%$ gain on mIOU. On the other hand, DPC variation performs $+0.6\%$ more accurate than the Mobilenet V2 with basic heads, which has $1.54$ times more GFLOPs.

ShuffleNet V2 with DPC heads outperforms state-of-art efficient networks on the Cityscapes test set. Table \ref{tbl:class_level} shows that we make a gain of 7.27\% mIOU over the best performing ENet architecture. Furthermore, on each of the classes highlighted here, ShuffleNet V2$+$DPC shows a great improvement. Also, as seen in the Table \ref{tbl:classcat}, our approach outperforms the previous methods on the category mIOU and instance level metrics.

\begin{figure}[t]
    \centering
    \includegraphics[width=.244\linewidth]{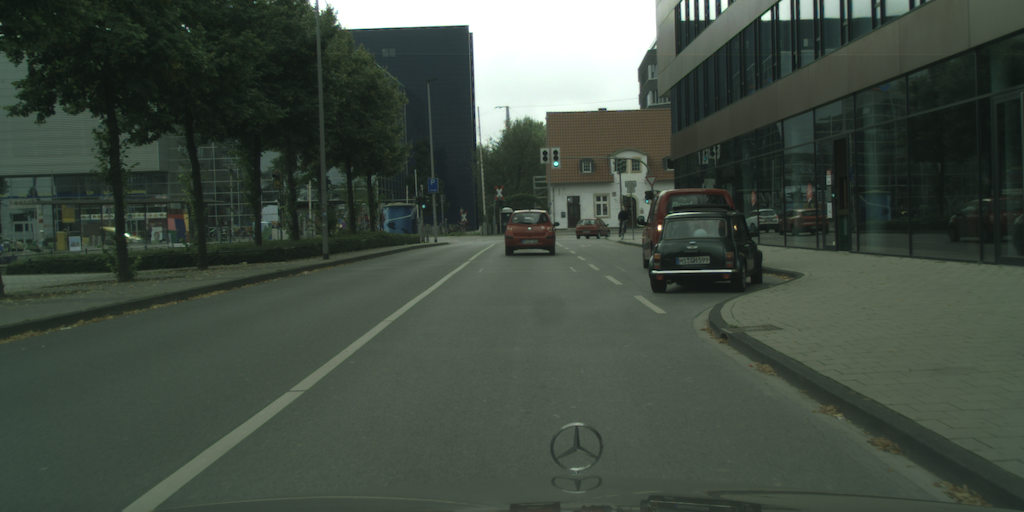}\hfill
    \includegraphics[width=.244\linewidth]{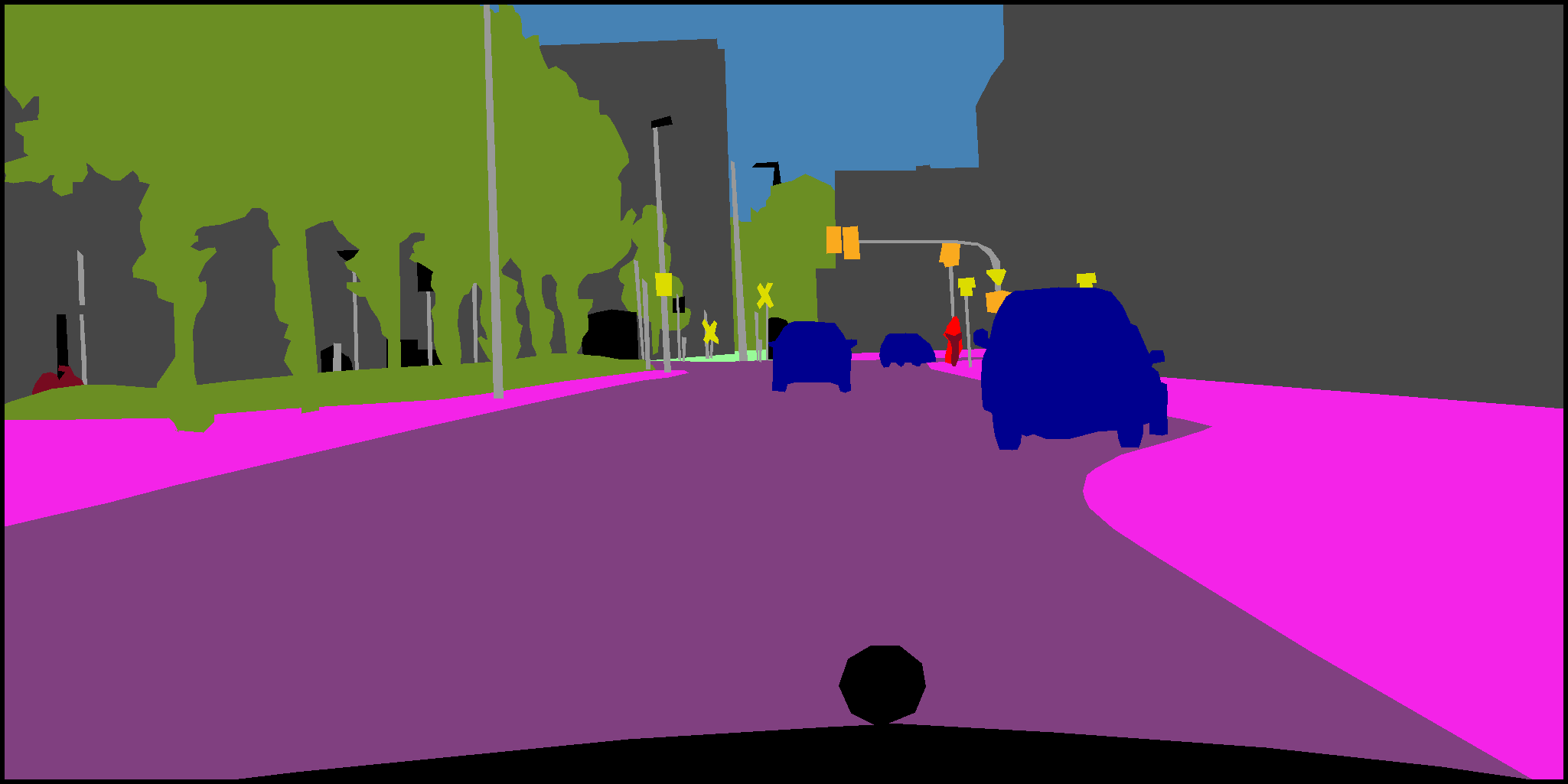}\hfill
    \includegraphics[width=.244\linewidth]{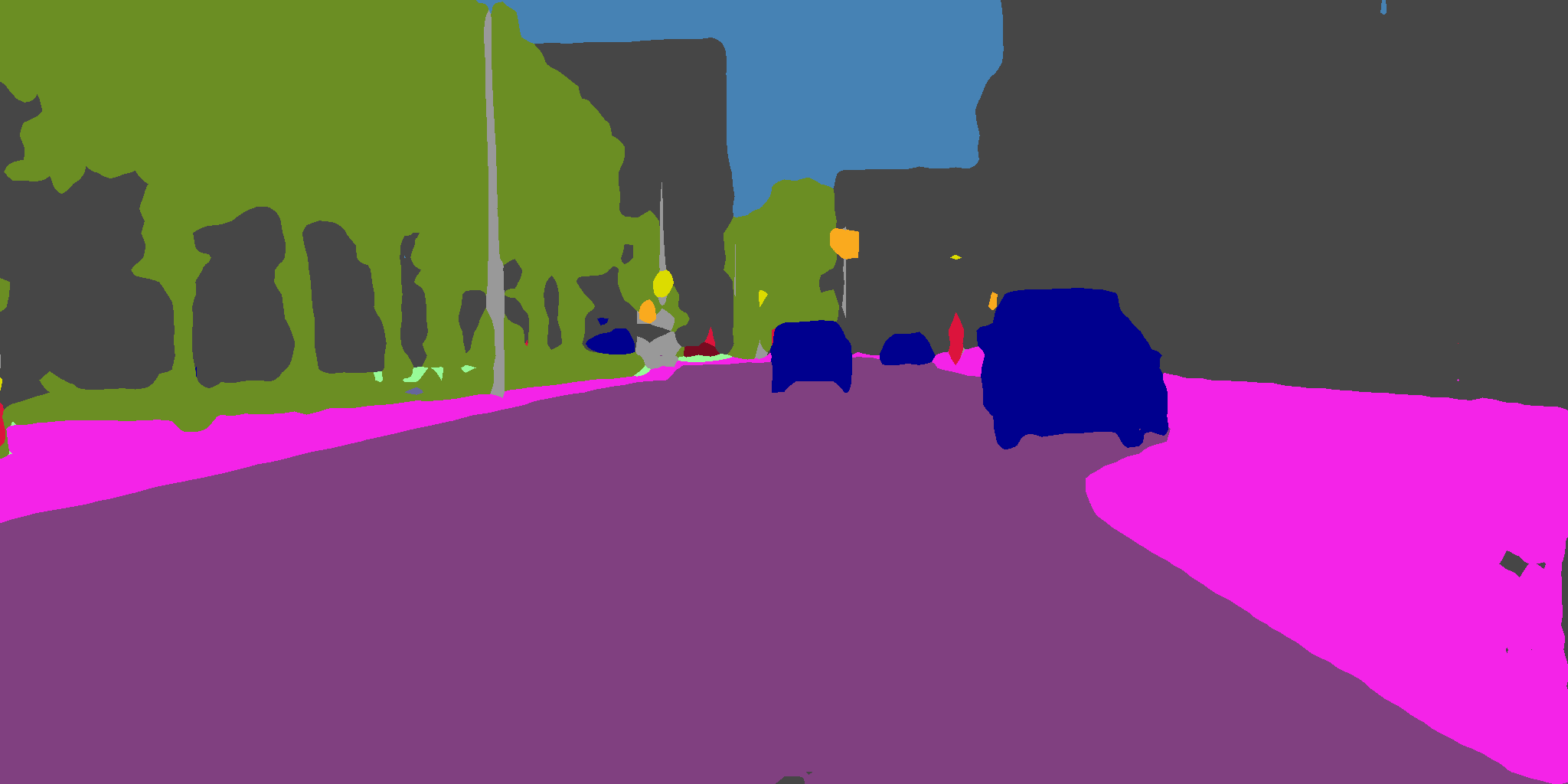}\hfill
    \includegraphics[width=.244\linewidth]{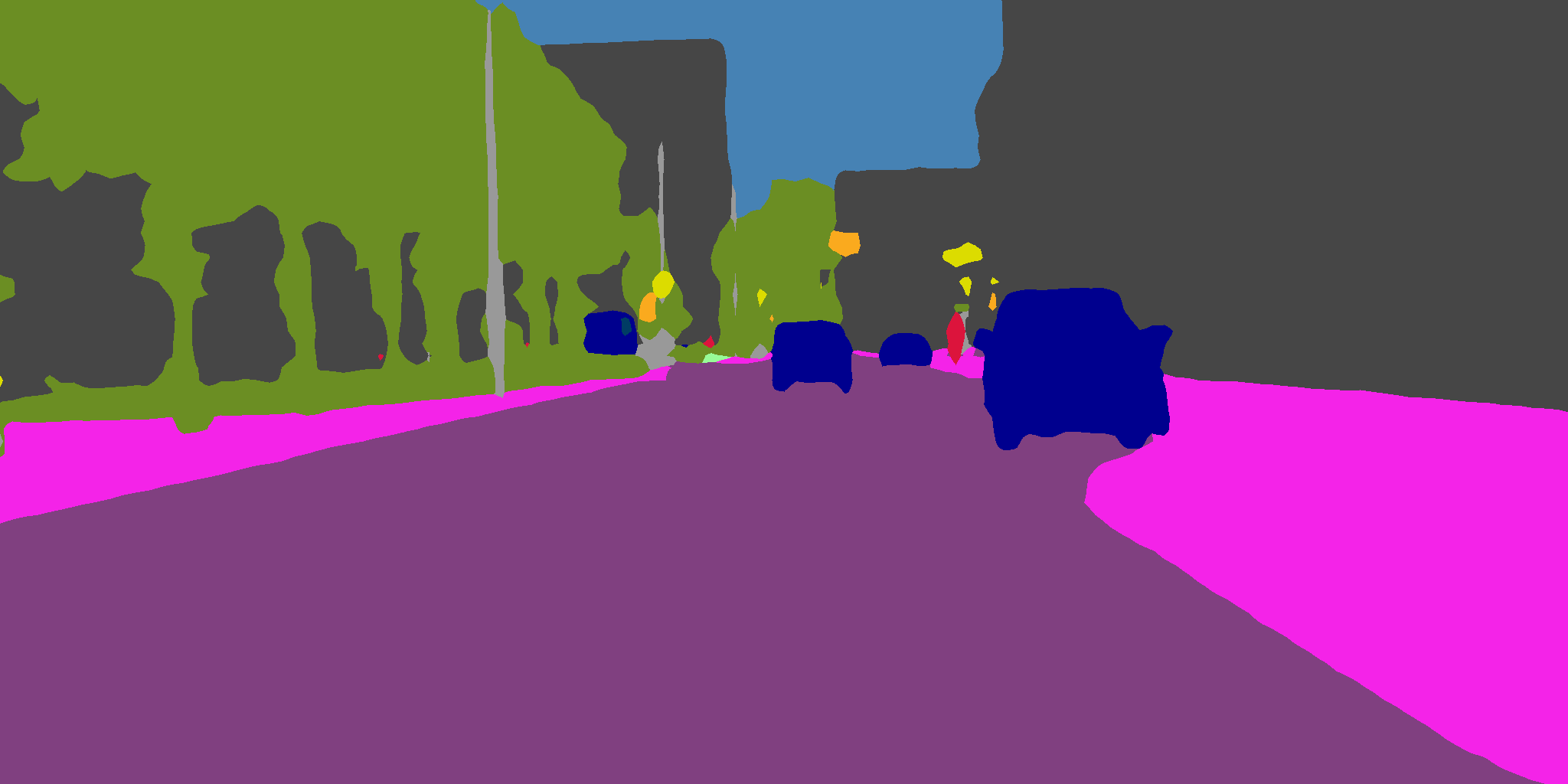}\\[0.5ex]
    \includegraphics[width=.244\linewidth]{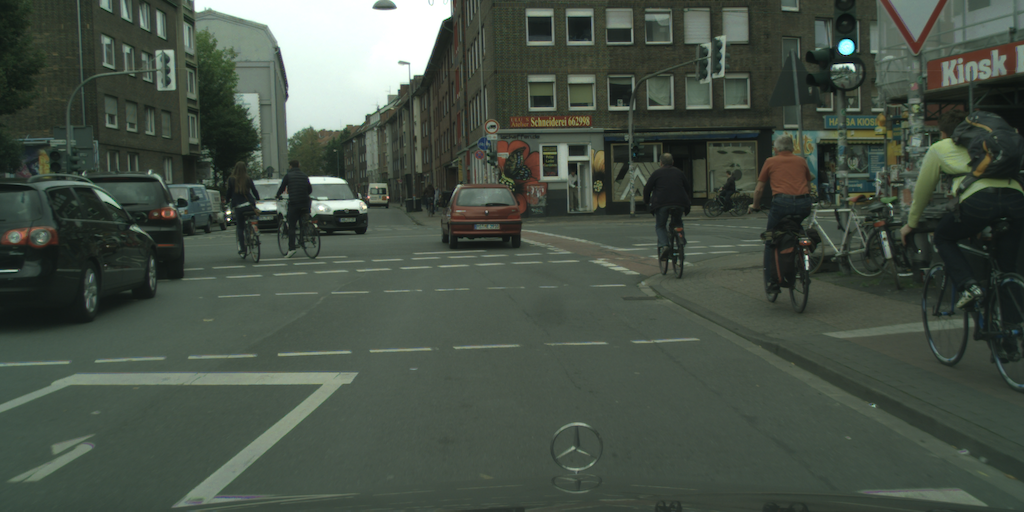}\hfill
    \includegraphics[width=.244\linewidth]{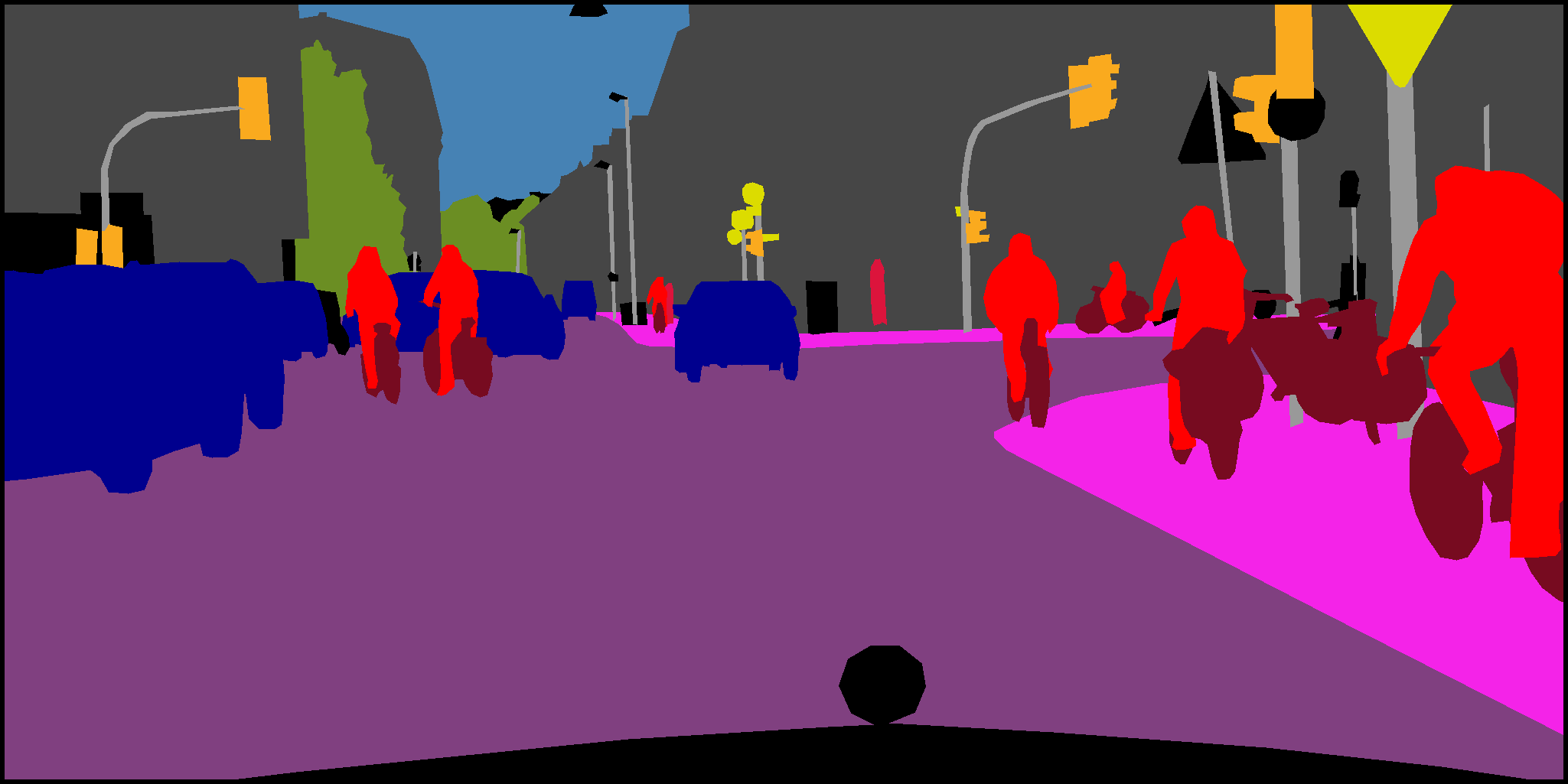}\hfill
    \includegraphics[width=.244\linewidth]{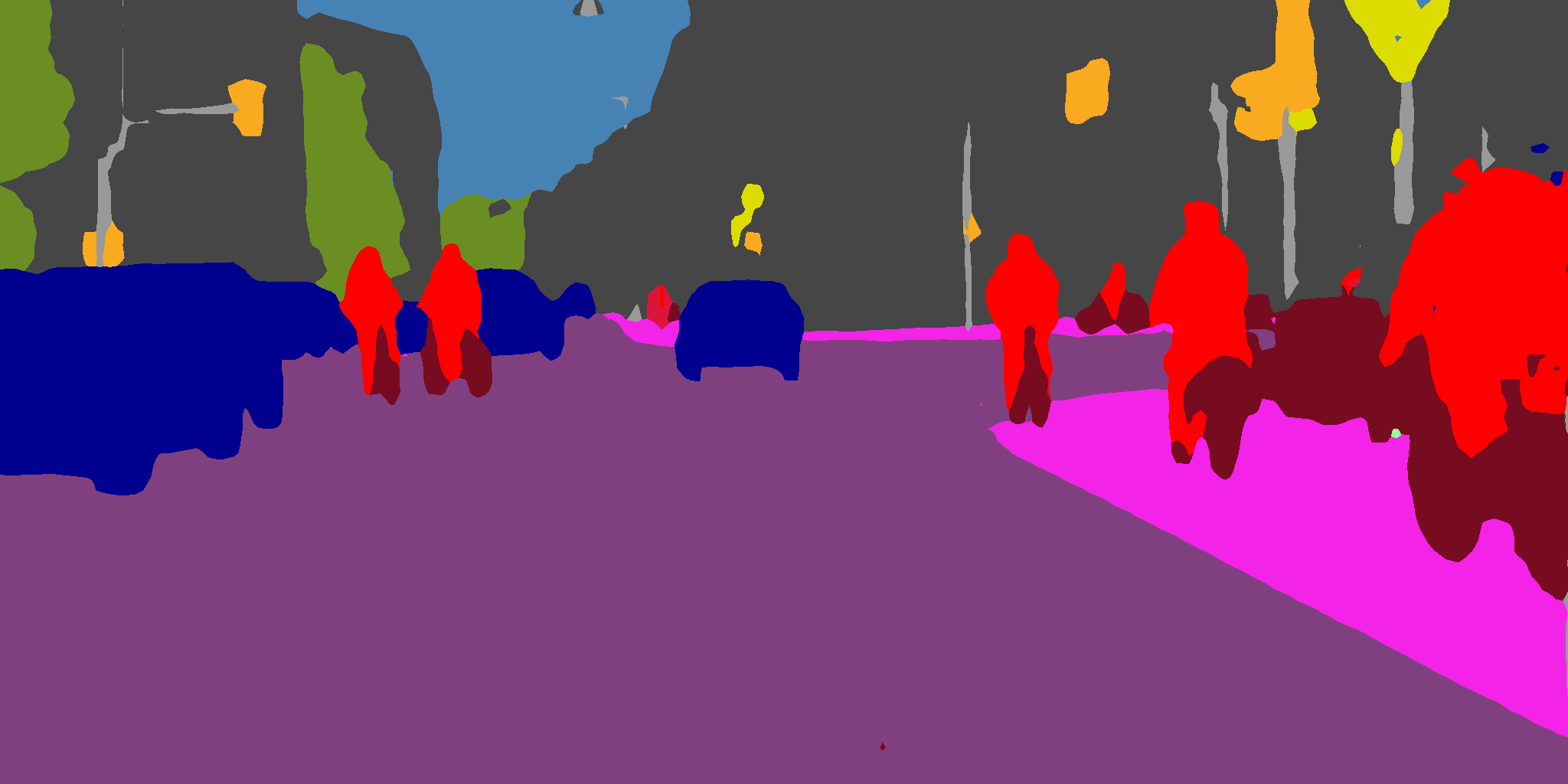}\hfill
    \includegraphics[width=.244\linewidth]{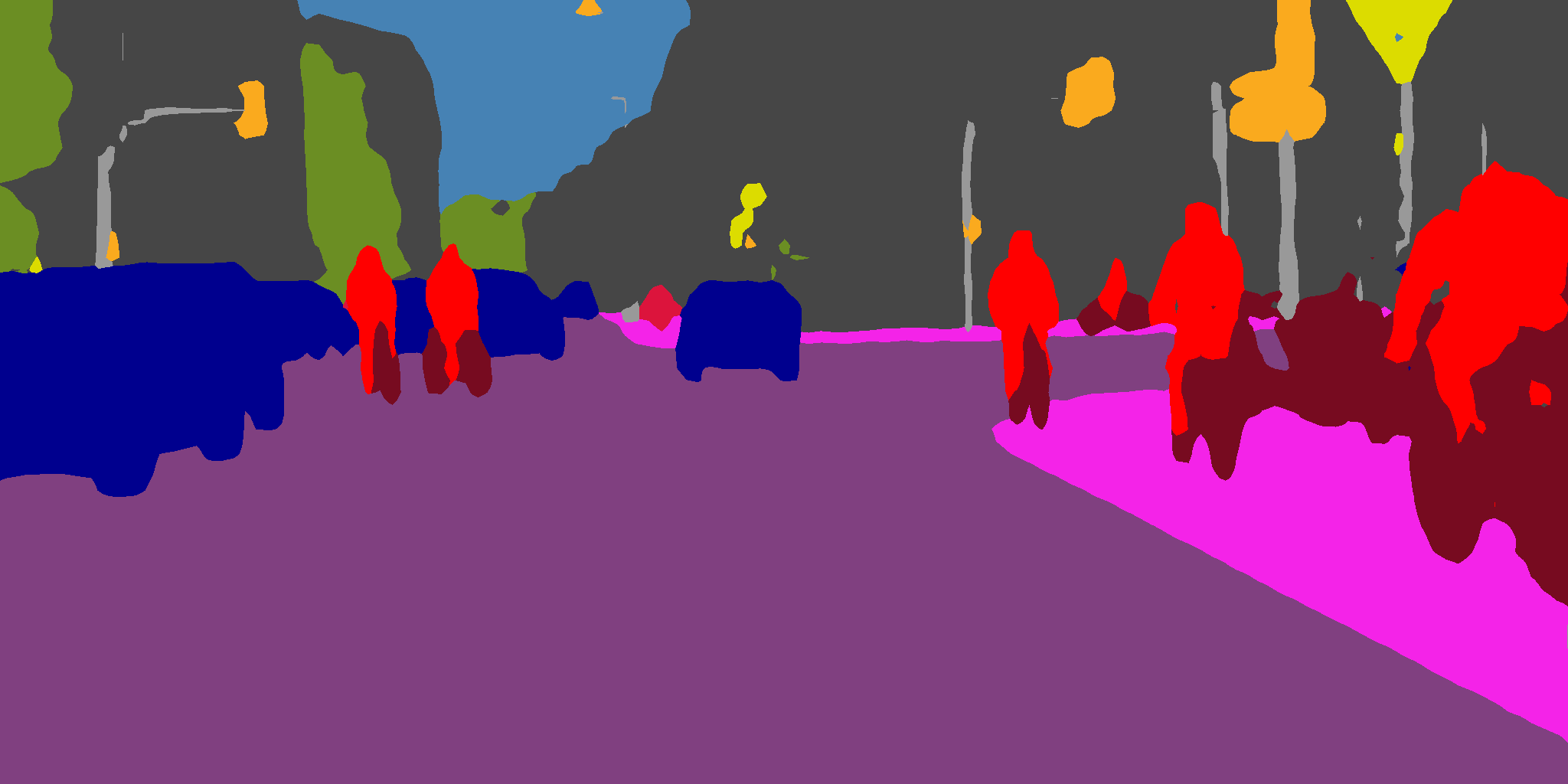}\\[0.5ex]
    \includegraphics[width=.244\linewidth]{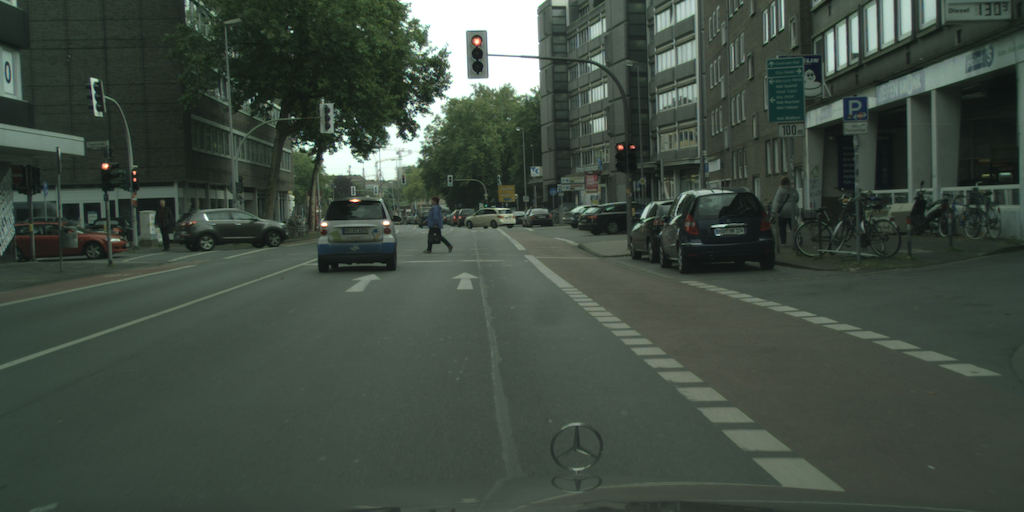}\hfill
    \includegraphics[width=.244\linewidth]{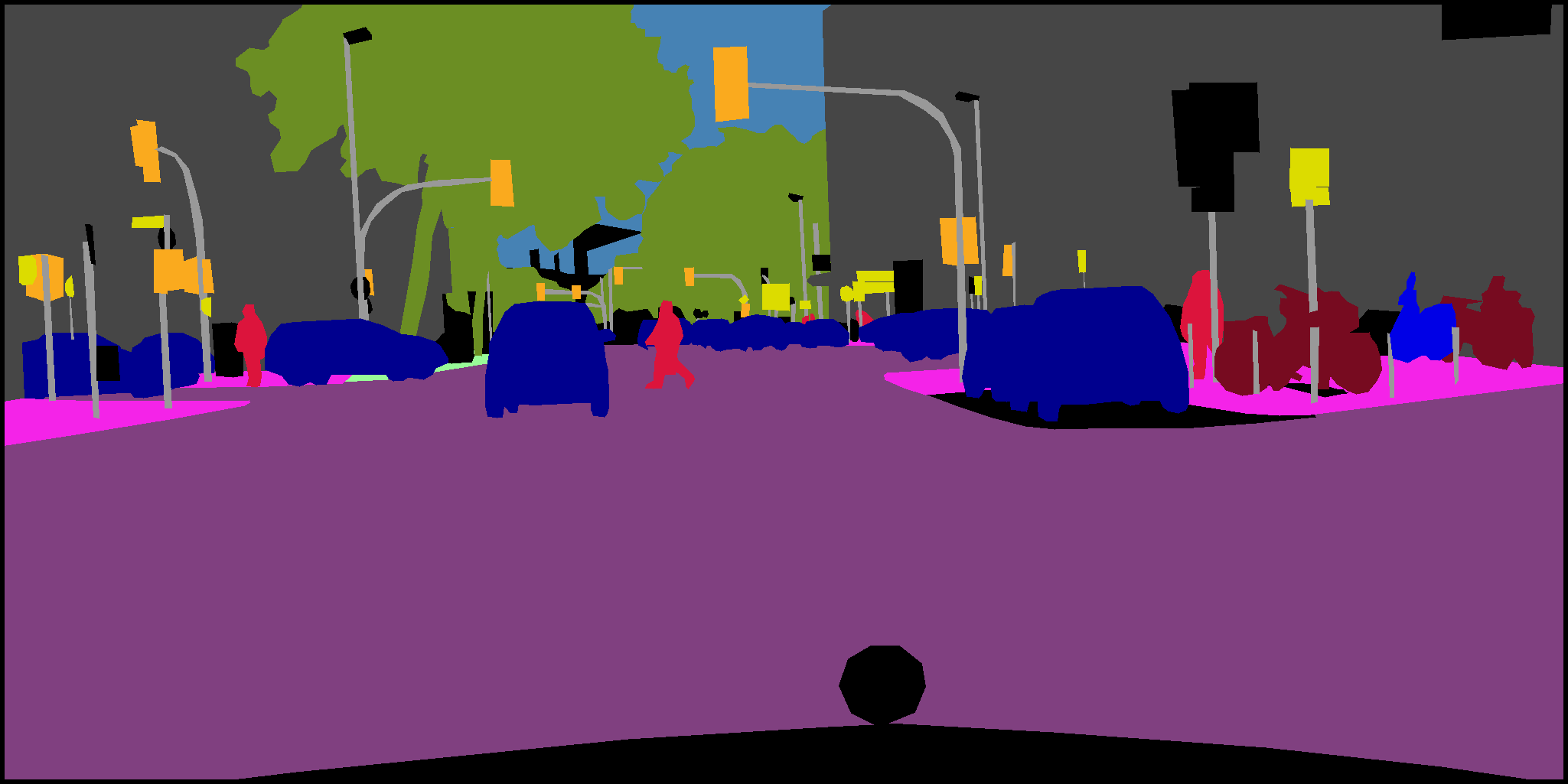}\hfill
    \includegraphics[width=.244\linewidth]{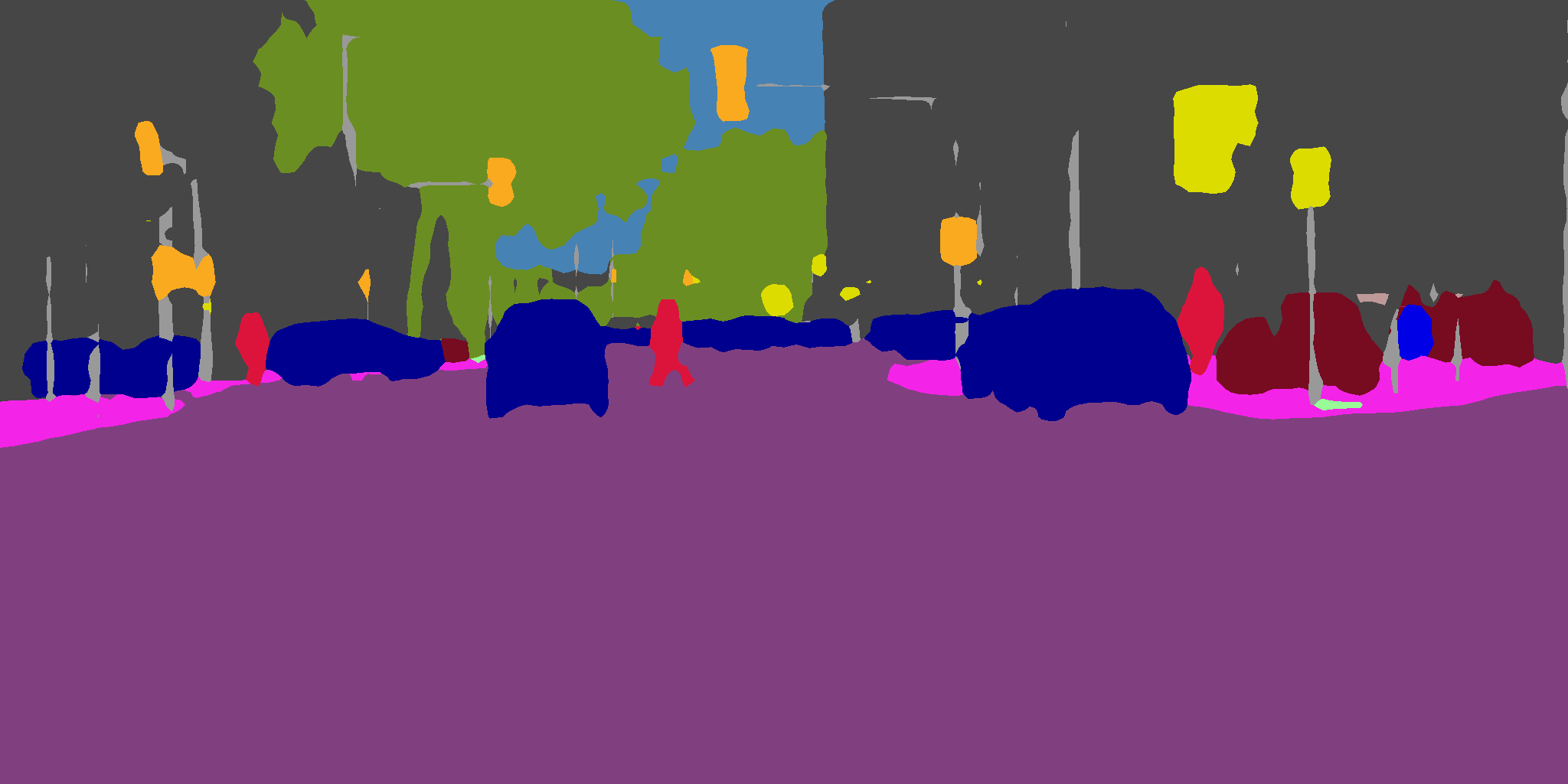}\hfill
    \includegraphics[width=.244\linewidth]{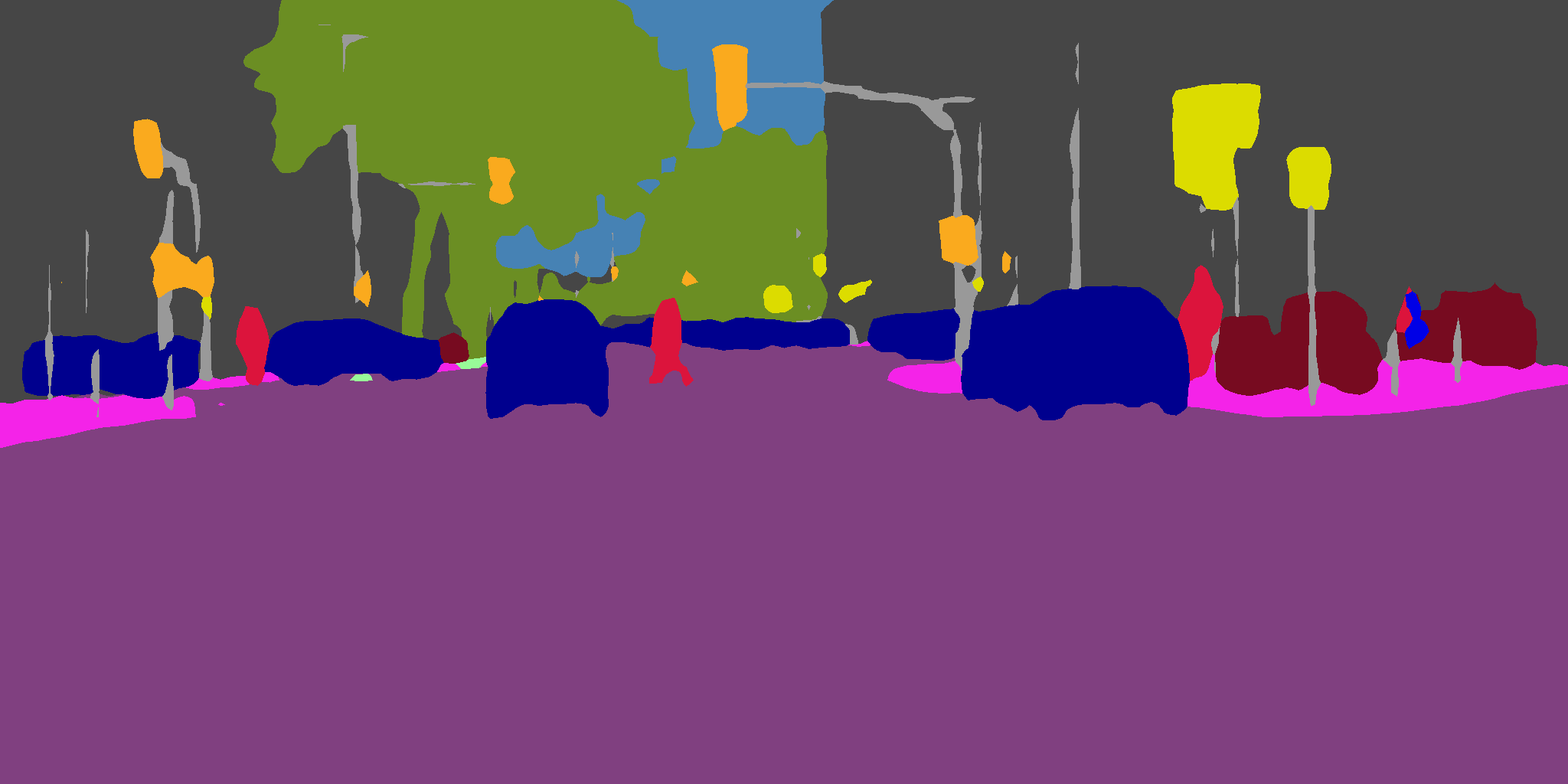}\\[-2ex]
    \subfloat[Input image]{\includegraphics[width=.244\linewidth]{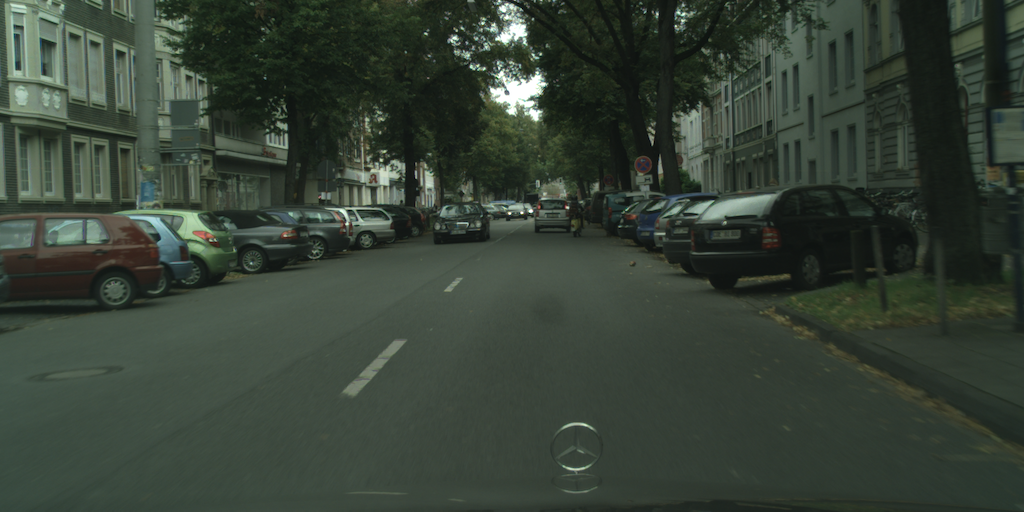}}\hfill
    \subfloat[Ground truth]{\includegraphics[width=.244\linewidth]{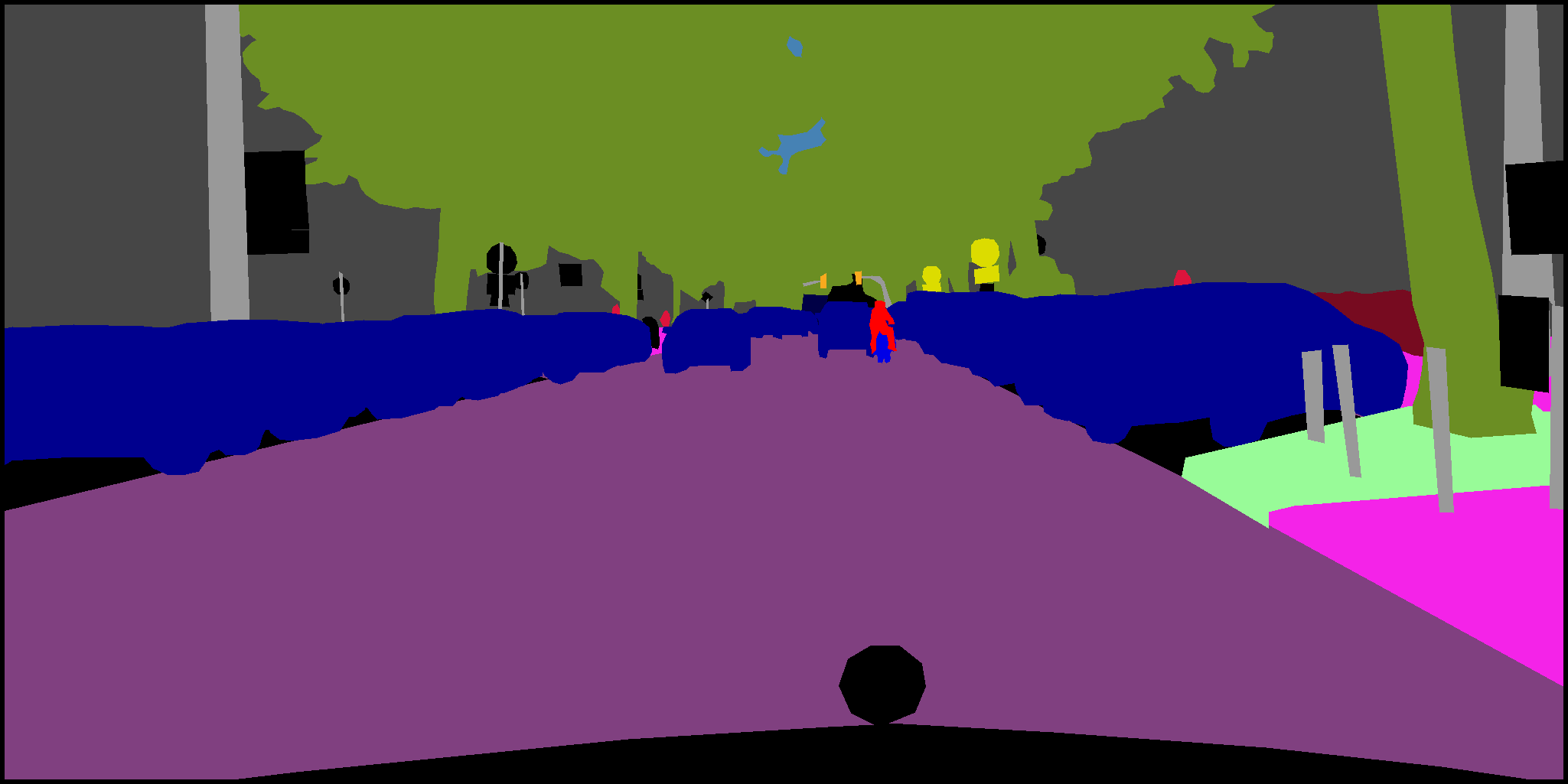}}\hfill
    \subfloat[Basic head]{\includegraphics[width=.244\linewidth]{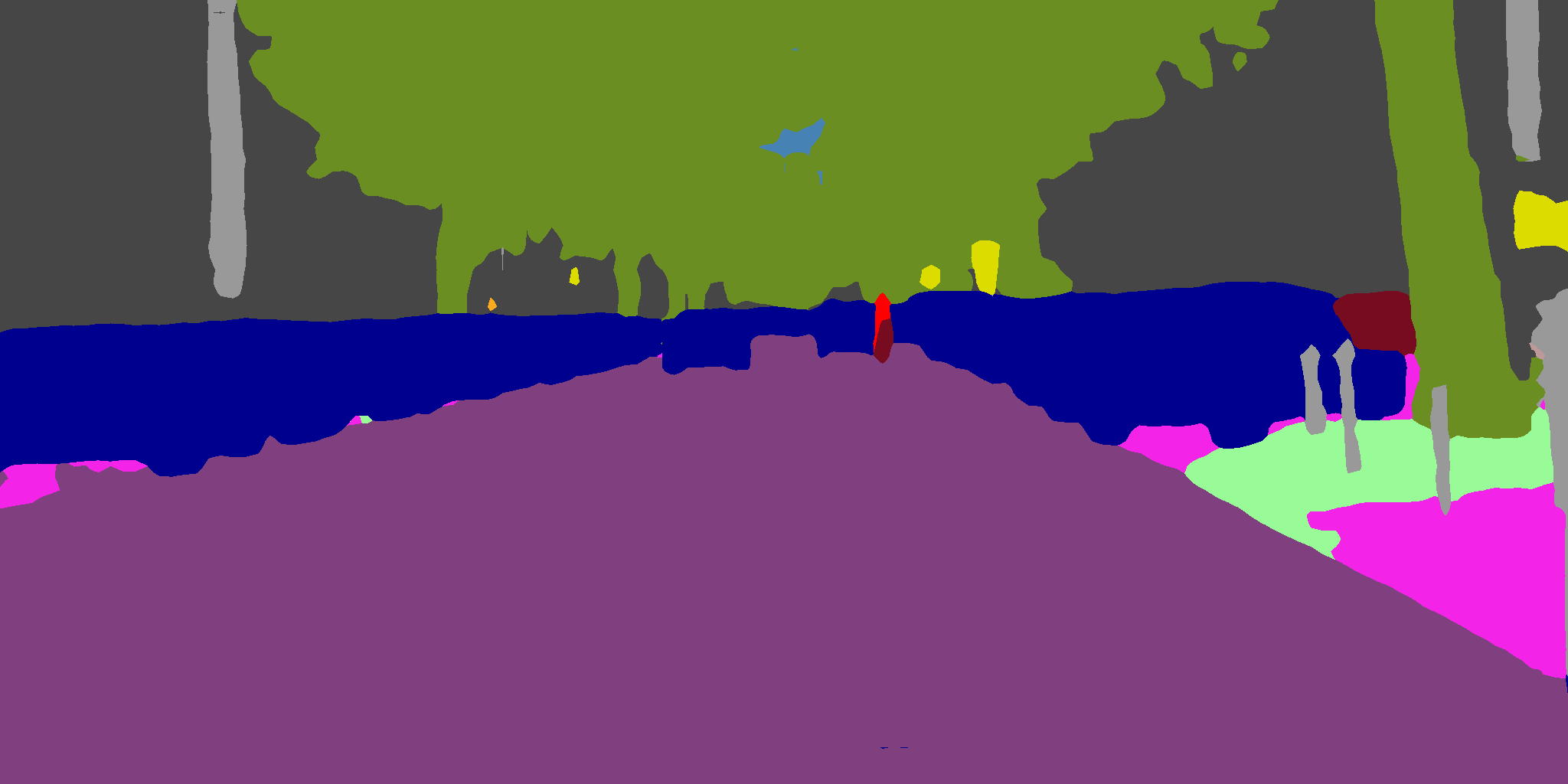}}\hfill
    \subfloat[DPC]{\includegraphics[width=.244\linewidth]{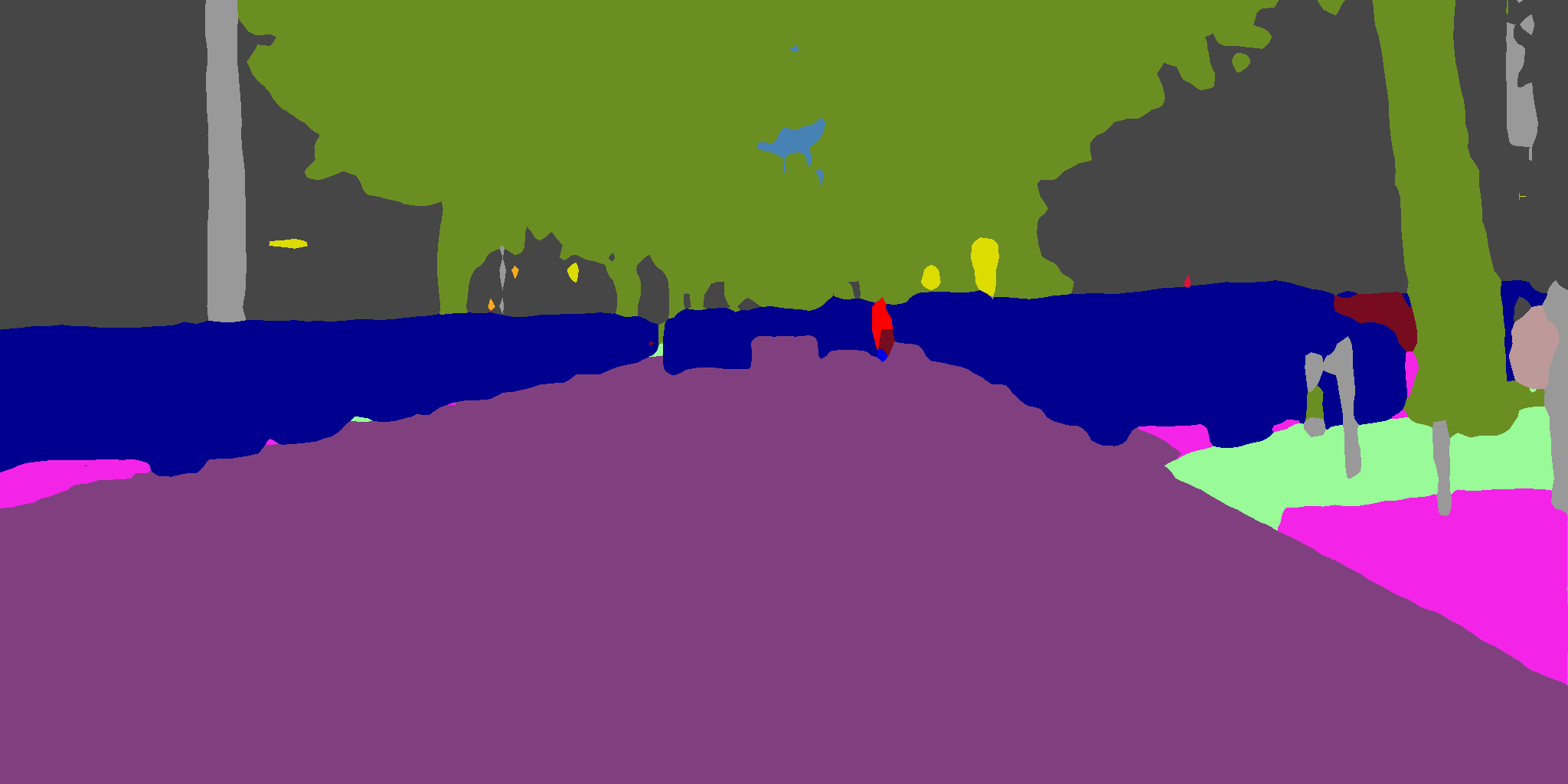}}
    \caption{Segmentation visualisations on the validation set. Best viewed in colour. (Black coloured regions on ground truth are ignored)}
    \label{fig:visuals}
\end{figure}

We visualize the segmentation masks generated both by the basic and DPC approach in Figure \ref{fig:visuals}. From the visuals, we can see that the DPC head variant is able to segment thinner objects such as poles better than the basic encoder head. The figure is best viewed in colour.

\begin{table}[t]
    \centering
    \caption{Comparison of class and category level accuracy on the test set.}
    \label{tbl:class_level}
    \resizebox{\linewidth}{!}{
        \begin{tabular}{lccccccccccc}
            Method                  & mIOU  & Building & Sky   & Car   & Sign  & Road  & Person & Fence & Pole  & Sidewalk & Bicycle \\
            \toprule
            SkipNet-MobileNet       & 61.52 & 86.19    & 92.89 & 89.88 & 54.34 & 95.82 & 69.25  & 39.40 & 44.53 & 73.93    & 58.15   \\
            Enet Lovasz             & 63.06 & 87.22    & 92.74 & 91.01 & 58.06 & 97.27 & 71.35  & 38.99 & 48.53 & 77.20    & 59.80   \\
            \hline
            ShuffleNet V2+DPC(Ours) & 70.33 & 90.7     & 93.86 & 93.95 & 66.93 & 98.11 & 78.47  & 50.93 & 51.47 & 82.46    & 67.48   \\
            \bottomrule
        \end{tabular}
    }
\end{table}

\begin{table}[t]
    \centering
    \caption{Comparison of class level accuracies efficient architectures on Cityscapes \textbf{test} set.}
    \label{tbl:classcat}
    \begin{tabular}{lccccc}
        Method                                      & Class IOU & Class iIOU & Cat. IOU & Cat. iIOU \\
        \toprule
        SegNet\cite{badrinarayanan2015segnet}       & 56.1      & 34.2       & 79.8     & 66.4      \\
        ShuffleSeg\cite{gamal2018shuffleseg}        & 58.3      & 32.4       & 80.2     & 62.6      \\
        SkipNet-MobileNet\cite{siam2018comparative} & 61.52     & 35.16      & 82.00    & 63.03     \\
        Enet Lovasz\cite{berman2018lovasz}          & 63.06     & 34.06      & 83.58    & 61.05     \\
        \hline
        ShuffleNet V2+DPC (Ours)                    & 70.33     & 43.58      & 86.48    & 69.92     \\
        \bottomrule
    \end{tabular}
\end{table}

\subsection{Inference Speed on Mobile Phone}

In this section, we provide inference speed results and our procedure for the measurements. As discussed in ShuffleNet V2 \cite{ma2018shufflenet}, we believe that the computational efficiency of the network should be measured on the target devices rather than purely comparing by the FLOPs. Thus, we convert the competing networks to Tensorflow Lite, a binary model representation for inferencing on mobile phones, for the speed evaluation. Tensorflow Lite has a limited number of available operations in its current implementation, thus complex architectures such as \cite{he2017mask} cannot be converted without implementing the missing operations on the Tensorflow Lite. But, both Mobilenet V2 and ShuffleNet V2 are compatible and require no additional custom operations.

The measurements were done on OnePlus A6003, Snapdragon 845 CPU, 6GB RAM, 16 + 20 MP Dual Camera, Android version 9, OxygenOS version 9.0.3. The device is stabilized in place, put to airplane mode, background services are disabled, the battery is fully charged and kept plugged to the power cable. The rear camera output image is scaled so that the smaller dimension is 224, then the middle of the image is cropped to get an input image of $224 \times 224$. The downscaling and cropping is not included in the inference time shown in Table \ref{tbl:inference} however the preprocessing of the input image, according to Equation \ref{eq:zero_mean_unit_range}, and the final ArgMax is included in the inference time. Inference speed measurements, presented in Table \ref{tbl:inference}, are averaged over 300 frames after 30 seconds of an initial warm-up period.

Table \ref{tbl:inference} shows that our approach of using ShuffleNet V2 as a backbone is capable of performing real-time, close to 20Hz, semantic segmentation on a mobile phone. Even with the complex DPC architecture, ShuffleNet V2 backbone ends up with 1.54 times fewer GFLOPs over the Mobilenet V2 with basic encoder head. Furthermore, model size is significantly lower which is desired for embedded devices that might have limited memory or storage size. One surprising finding is that the Mobilenet V2 variants show much higher variance in the inference time compared to the ShuffleNet V2 backbone, thus our approach provides a more stable fps.

\begin{table}[t]
    \centering
    \caption{Inference performance on OnePlus A6 with an input size of $224 \times 224$.}
    \label{tbl:inference}
    \begin{tabular}{lccccccc}
        Backbone                       & Encoder & GFLOPs & Inference (ms) & Var (ms) & FPS   & Size (MB) \\
        \toprule
        \multirow{2}{*}{ShuffleNet V2} & Basic   & 0.47   & 50.89          & 0.57     & 19.65 & 4.6       \\
                                       & DPC     & 0.65   & 64.89          & 3.53     & 15.41 & 6         \\
        \hline
        \multirow{2}{*}{MobileNet V2}  & Basic   & 1.00   & 101.46         & 25.04    & 9.86  & 8.4       \\
                                       & DPC     & 1.18   & 116.16         & 44.01    & 8.61  & 9.9       \\
        \bottomrule
    \end{tabular}
\end{table}

\section{Conclusion}

In this paper, we have described an efficient solution for semantic segmentation by achieving state-of-art inference speed without compromising on the accuracy. Our approach shows that ShuffleNet V2 is a powerful and efficient backbone for the task of semantic segmentation. It achieves 70.33\% mIOU when combined with the DPC head, and 67.7\% mIOU\footnote{The validation set result.} combined with the basic encoder head on Cityscapes challenge. Furthermore, we showed that our network is capable of running real-time, the DPC head at 15.41 fps and the basic encoder head at 19.65 fps, on a mobile phone with an input image size of $224 \times 224$. Future work is to implement an efficient decoder architecture to get more refined edges along the borders of the objects on the segmentation mask.

\bibliographystyle{splncs04}
\bibliography{mybibliography}

\end{document}